\documentclass[nohyperref]{article} 
\usepackage{iclr2023_conference,times}

\usepackage{amsmath,amsfonts,bm}

\def\eqref#1{equation~\ref{#1}}

\def\1{\bm{1}}

\DeclareMathAlphabet{\mathsfit}{\encodingdefault}{\sfdefault}{m}{sl}
\SetMathAlphabet{\mathsfit}{bold}{\encodingdefault}{\sfdefault}{bx}{n}

\usepackage[hidelinks]{hyperref}
\definecolor{mydarkblue}{rgb}{0,0.08,0.45}
\hypersetup{colorlinks,linkcolor={mydarkblue},citecolor={mydarkblue},urlcolor={red}}
\usepackage{url}

\usepackage{amsmath}
\usepackage{amssymb}
\usepackage{mathtools}
\usepackage{amsthm}
\usepackage{algorithm}
\usepackage{algorithmic}
\usepackage{dsfont}  
\usepackage{makecell}
\usepackage{booktabs}

\usepackage{tikz}
\usetikzlibrary{shapes}
\usetikzlibrary{plotmarks}

\theoremstyle{plain}
\newtheorem{theorem}{Theorem}[section]

\theoremstyle{definition}

\newtheorem{assumption}[theorem]{Assumption}
\theoremstyle{remark}
\newtheorem{remark}[theorem]{Remark}

\newcommand{\cS}{\mathcal{S}}
\newcommand{\cD}{\mathcal{D}}
\newcommand{\cA}{\mathcal{A}}
\newcommand{\RR}{\mathbb{R}}

\newcommand{\PP}{\mathbb{P}}

\usepackage{natbib}
\setcitestyle{authoryear,round,citesep={,},aysep={},yysep={;}}

\title{A Tutorial: An Intuitive Explanation of Offline Reinforcement Learning Theory}

\author{Fengdi Che \\
Department of Computer Science\\
University of Alberta\\
\texttt{fengdi@ualberta.ca} \\
}

\iclrfinalcopy 
\begin{document}
\maketitle

\begin{abstract}
    Offline reinforcement learning (RL) aims to optimize the return given a fixed dataset of agent trajectories without additional interactions with the environment. While algorithm development has progressed rapidly, significant theoretical advances have also been made in understanding the fundamental challenges of offline RL. However, bridging these theoretical insights with practical algorithm design remains an ongoing challenge. In this survey, we explore key intuitions derived from theoretical work and their implications for offline RL algorithms.
    
    We begin by listing the conditions needed for the proofs, including function representation and data coverage assumptions. 
    Function representation conditions tell us what to expect for generalization, and data coverage assumptions describe the quality requirement of the data.
    We then examine counterexamples, where offline RL is not solvable without an impractically large amount of data. 
    These cases highlight what cannot be achieved for all algorithms and the inherent hardness of offline RL.
    Building on techniques to mitigate these challenges, we discuss the conditions that are sufficient for offline RL. 
    These conditions are not merely assumptions for theoretical proofs, but they also reveal the limitations of these algorithms and remind us to search for novel solutions when the conditions cannot be satisfied. \looseness=-1
    
\end{abstract}

\section{Introduction}
In recent years, offline or batch reinforcement learning (RL) has gained increasing attention \citep{levine2020offline}. 
Offline RL focuses on learning from pre-collected datasets without additional interactions with the environment. 
This approach is particularly valuable in applications where the execution of an unqualified policy is risky or expensive, such as healthcare, autonomous driving, and recommender systems \citep{zhao2018deep}.
At the same time, the absence of online interaction poses unique challenges: partial coverage of the data, distribution shifts, and function approximation errors often complicate the learning process.

This survey centers on offline policy optimization, where the primary objective is to learn a near-optimal policy. 
Our chief concern is statistical efficiency, namely, how many samples are necessary to achieve a given level of performance, while we do not address computational complexity in this work.
This focus aligns with the broader question of tractability: if a task cannot be solved with a reasonable amount of data, then analyzing its computational requirements becomes unnecessary.

The goal of this survey is to present theoretical results in a way that is accessible to non-theory researchers, highlighting how offline RL practitioners can benefit from these insights in several ways. Our goal is to understand when offline RL is solvable and what assumptions are necessary to ensure tractability. Given the practical demands of large-scale problems, we focus on scenarios that employ function approximation \citep{sutton1998introduction}. Designing an effective offline RL algorithm requires addressing fundamental questions about function representation, such as how large the neural network should be and the quality of the dataset. To formalize these considerations, we introduce assumptions on function representation and data coverage in Section \ref{sec:assump}.

By examining counterexamples where offline RL is intractable, theoretical insights help clarify the intrinsic hardness of the offline RL setting \citep{zanette2021exponential,foster2022offline}. These insights guide the development of truly effective techniques rather than relying on brute-force approaches. The assumptions that suffice for supervised or online learning \citep{wang2021statistical, weisz2023online} do not necessarily guarantee polynomial sample complexity in offline policy optimization. Offline learning presents additional challenges, many of which are well-characterized by the lower-bound results discussed in Section \ref{sec:hardness}.

Next, we analyze key challenges arising from bootstrapping bias, over-estimated extrapolation, insufficient coverage, and spurious data, emphasizing why common techniques such as $\lambda$-return \citep{sutton1989implementation} and pessimism \citep{kumar2020conservative} are needed to mitigate these issues. 
Building on these techniques, we discuss the sufficient conditions \citep{golowich2024role} under which algorithms succeed. In practice, understanding the limitations of our algorithms is crucial, allowing us to avoid endless model tuning and seeking extra solutions. 
For example, we should not expect to solve inherently NP-hard tasks within polynomial time.
When tackling more complex scenarios without meeting sufficient conditions, novel techniques are often required to address challenges arising from assumption failures.

Finally, we briefly discuss the low-adaptive setting, which lies between online incremental learning and offline learning. In this setting, a batch of data is collected after each policy update, allowing for fewer policy switches relative to the total number of samples used. While it is theoretically feasible, what challenges prevent us from designing a low-adaptive algorithm?

\paragraph{Paper organization.} We first introduce the fundamental questions \citep{lattimore2020bandit} that theory researchers are investigating in Section \ref{sec:key_questions}. Then, in Section \ref{sec:math}, we establish the mathematical notations and formally define the problems of interest. Common assumptions are presented in Section \ref{sec:assump}. In Sections \ref{sec:hardness} and \ref{sec:sufficient}, we discuss inherent challenges and sufficient conditions, respectively. For a first read, it's fine to focus on the introduction to theoretical questions and then jump to the key takeaways at the end of each section. 

\section{An Introduction to Theoretical Questions}
\label{sec:key_questions}
Theory researchers often focus on two fundamental questions about offline RL tasks: first, under what conditions can these tasks be solved, and second, how efficiently can tasks be solved under those conditions? For the first question, the usual approach is to identify the minimum sufficient assumptions that guarantee tractability on a family of Markov Decision Processes (MDPs) and datasets. As illustrated in the figure below, researchers start with broad assumptions and then identify counterexamples which are not tractable. Finding a counterexample means that the set of tasks under these assumptions cannot all be solved using polynomial samples.

To address these weaknesses of each counterexample, they introduce stronger assumptions. As shown in the figure, the stronger the assumptions, the smaller the set of solvable tasks. Once an algorithm is developed to solve all tasks within the set defined by these stronger conditions, a sufficient set of assumptions is established. The focus then shifts toward gradually weakening these assumptions until a new counterexample emerges. This iterative process continues until the minimal set of sufficient assumptions is identified.

\begin{figure}[ht]
\begin{center}
\begin{tikzpicture}
    
    \draw[thick] (0,0) ellipse (6 and 3);
    \draw[thick] (0,0) ellipse (5 and 2.5);
    \draw[thick] (0,0) ellipse (2 and 1);

    \filldraw (-4,0) circle (3pt);
    \node[above right] at (-4.7,-0.8) {Counterexample};
    
    
    \node[above right] at (-1,2.5) {All Tasks};
    \node[above right] at (-3,1.2) {Tasks Satisfying the Broader Assumptions};
    \node[above right] at (-1.5,-0.2) {Stronger Conditions};    
\end{tikzpicture}
\end{center}
\end{figure}

Once there is consensus on assumptions that enable learning, attention shifts to efficiency.
The worst-case upper bound states that a particular algorithm can solve any task satisfying the conditions with no more than some specified number of samples. The upper bound specifies the maximum amount of data required to solve any task in the set.
The name "worst-case" comes from the goal of solving any task in the set, even the worst case.
The worst-case lower bound explains that, regardless of which algorithm is used, there will always be an “unsolvable” counterexample if fewer samples are available. 
The low bound specifies the minimum amount of data required to solve any task in the set.

How can we assess the quality of upper and lower bounds? Theorists aim to close the gap between them by constructing harder counterexamples to increase the lower bound and designing more efficient algorithms to decrease the upper bound.
When the upper-bound complexity matches the lower bound, the bounds cannot be further improved. 
An algorithm that achieves this match is called minimax optimal. 
Reaching this match implies that no algorithm can solve the problem with fewer samples, meaning the algorithm has attained the best possible efficiency under the given assumptions. \looseness=-1

\section{Background}
\label{sec:math}
\paragraph{Notation} 
We let $\Delta(\mathcal{X})$ denote the set of probability distributions over a finite set $\mathcal{X}$. 
Let $\RR$ denote the set of real numbers.
Let $<\cdot,\cdot>$ denote the inner product between vectors and $\|\cdot\|$ represent the corresponding norm.
Let $[H]$ denote the set of integers counting from one to $H$.

\paragraph{Markov Decision Process}
We consider finite-horizon Markov decision processes (MDP) \citep{sutton1998introduction, puterman2014markov}
\(
M= (\mathcal{S}, \mathcal{A}, H, \mathcal{P}, r)
\)
with a finite state space \(\mathcal{S}\), an action space \(\mathcal{A}\), the horizon \(H\), the transition probability 
\(\mathcal{P} = \{ \mathcal{P}_h \}_{h=1}^H\), and reward functions \(r = \{ r_h \}_{h=1}^H\).
We assume each reward is bounded, i.e., \(r_h \in [0,1]\) for all \(h \in [H]\) and assume a fixed initial state denoted by $x$ for simplicity.

A \emph{policy} 
$\pi : \mathcal{S} \to \Delta(\mathcal{A})$ 
describes which action to take at each state. 
The agent starts at an initial state $s_1=x$.
At each step $i$, the agent takes an action \(a_i \sim \pi_i(\cdot \mid s_i)\) at state \(s_i\), receives a reward $r_i(s_i,a_i)$ and transits to the next state 
\(s_{i+1} \sim \mathcal{P}_i(\cdot \mid s_i, a_i)\). Repeating $H$ steps, the agent reaches the termination state, and the trajectory ends. 

The \emph{action-value function} or Q-value function, denoted by \(q_h^\pi: \mathcal{S} \times \mathcal{A} \to \mathbb{R}\) at step \(h \in [H]\),  for a policy $\pi$ represents the cumulative return gained starting from any state-action pair at step $h$ and is defined as
\begin{equation}
    q_h^\pi(s, a)
    \;=\;
    \mathbb{E}_\pi \Bigl[
        \sum_{i=h}^H 
            r_i(s_i, a_i)
        \,\Bigm|\,
        s_h = s,\,
        a_h = a
    \Bigr].
    \label{eq:action-value}
\end{equation}
\(\mathbb{E}_\pi\) represents taking expectation over trajectories induced by \(\pi\). 
The \emph{state-value function}, denoted by \(v_h^\pi : \mathcal{S} \to \mathbb{R}\) at step \(h \in [H]\),  for a policy $\pi$ is defined as
\begin{equation}
    v_h^\pi(s)
    \;=\;
    \sum_{a\in\cA} \pi_h(a|s)q_h^\pi(s,a).
    \label{eq:state-value}
\end{equation}
Due to the bounded rewards, all value functions are also bounded, denoted by $V_\textrm{max}.$ 

Under certain conditions \citep{puterman2014markov}, there exists an optimal policy 
$\pi^\star$ which gives the highest value and action-value functions among all policies, defined as: for all states $s$,
\[
  v_h^\star(s) 
  \;=\; v_h^{\pi^\star}(s) 
  \;=\; 
  \sup_{\pi} 
  \bigl\{ 
    v_h^\pi(s) 
  \bigr\},
  \quad
  q_h^\star(s, a) 
  \;=\; 
  q_h^{\pi^\star}(s, a) 
  \;=\; 
  \sup_{\pi} 
  \bigl\{ 
    q_h^\pi(s, a) 
  \bigr\}.
\]

We define the \emph{optimal Bellman operator} $\mathcal{T}_h$ acting on a function $q_{h+1}$ 
at step $h$ by
\[
  \mathcal{T}_h(q_{h+1})(s, a)
  \;=\;
  r_h(s, a)
  \;+\;
  \mathbb{E}_{s' \sim \mathcal{P}_h(\cdot \mid s,a)}
  \Bigl[
    \max_{a' \in \mathcal{A}}
    q_{h+1}(s', a')
  \Bigr].
\]

\paragraph{Linear Function Approximation}
Function approximation is commonly employed when the state space is extremely large or continuous. In this context, we restrict ourselves to a class of candidate value functions,
    \(
      \mathcal{F} \;\subset\; 
      \Bigl(\mathcal{S} \times \mathcal{A} \,\to\, [0, V_{\max}]\Bigr)
    \),
which are used to approximate value functions. The estimated Q-values are then denoted by $\hat{q} \in \mathcal{F}$. 

In this work, we focus on linear function approximation. The value function at each step $h$ is estimated as
$q^\pi_h(s_h, a) \approx \phi(s_h, a)^\top\theta_h$, 
where $\theta_h\in\RR^d$ is a parameter vector, and $\phi:\cS\times\cA\rightarrow\RR^d$ is a given feature representation. 
Also, the feature vector is usually assumed to have bounded norm, that is, for all $(s,a) \in \cS \times \cA$,
\(\|\phi(s,a)\|_2 \leq 1\).

\paragraph{Offline Policy Optimization}
In offline learning, a dataset $\mathcal{D}$ of transition tuples 
$\{(s_h^j, a_h^j, r_h^j, (s'_{h+1})^j)\}_{h=1,j=1}^{H,K}$ is pre-collected, where the reward and next state are sampled from an underlying MDP. Given a transition $(s_h,a_h,r_h,s'_{h+1})$, the one-step bootstrapping target for $\hat{q}(s_h,a_h)$ is an approximate of the optimal Bellman operator, equaling to $r_h^j + \max_{a'} \hat{q}(s'_{h+1},a')$.

There are two primary ways to collect the dataset in offline RL. Oracle-generated data is collected from a fixed distribution $\mu=\{\mu_h\}_{h=1}^H$ independently, where $(s_h,a_h) \sim \mu_h$ and the next state is sampled as $s'_{h+1} \sim \mathcal{P}_h(\cdot|s_h,a_h)$. In this approach, transitions are independent, meaning any state in the state space can be sampled without needing to be reachable from the previous step's state.
Policy-induced data is collected by executing a behaviour policy, resulting in a dataset that consists of complete trajectories. 

An offline task is defined by three factors $\langle M, \cD,\mathcal{F} \rangle$ and the goal is to learn a policy \(\hat{\pi}\) that maximizes the expected cumulative reward. 
Accordingly, we define the performance metric as
\begin{equation}
    \mathrm{SubOpt}\bigl(\hat{\pi}; x\bigr)
    \;=\;
    v_1^*(x)
    \;-\;
    v_1^{\hat{\pi}}(x),
\end{equation}
which measures the suboptimality of policy \(\hat{\pi}\) given the initial state \( s_1=x\).

\section{Common Assumptions}
\label{sec:assump}
In offline learning, a pre-collected dataset is given for all training, and no further interactions with the environment are allowed. 
Thus, this inherent constraint places a fundamental limit on the best possible policy an algorithm can learn. 
On the other hand, even though the dataset does not give us full generalization to know the best policy, a certain amount of generalization is needed to avoid visiting all state-action pairs. 
The generalization requires a well-chosen representation. 
Therefore, many theoretical results in offline RL hinge on two core elements: coverage assumptions, which dictate how well the dataset spans the space of relevant state-action pairs, and representation assumptions, which ensure the learning target is within the capacity of the model.
In this section, we will present common coverage and representation assumptions in offline RL research. 

\subsection{Representation}
Our goal is to learn the optimal policy, determining which action yields the highest return in each state. One approach is to first estimate action-value functions and then derive the policy accordingly. For instance, the learned policy can be expressed as $\hat{\pi}(s) = \arg\max_{a\in\cA} \hat{q}(s,a)$.
To ensure accurate value estimation, we would expect that the value functions of all policies lie within the chosen function class. In the following, we focus on linear function approximation as a concrete realization of these representation assumptions.

\begin{assumption}[All-Policy Value Function Realizability]
\label{assump:all-realizability}
For every policy $\pi$, there exist parameters 
$\theta_{1}^\pi, \dots, \theta_{H}^\pi \in \mathbb{R}^d$ such that for all $(s, a) \in \mathcal{S} \times \mathcal{A}$ 
and for each $h \in [H]$,
\[
q_h^\pi(s, a) \;=\; \phi(s, a)^\top \theta_h^\pi,
\]
or $q_h^\pi \in \mathcal{F}$ for a general function class.
\end{assumption}

While such realizability assumptions may suffice in supervised learning, reinforcement learning relies on bootstrapping using estimated values rather than directly observed targets, causing estimation biases to compound over multiple steps. To address this, many analyses impose additional requirements, often referred to as Bellman completeness. It ensures that the bootstrapping target, or the application of the Bellman operator to any function in the class, remains within the representational scope of the function class.

\begin{assumption}[Linear Bellman Completeness]
\label{assump:low-inherent-bellman-error}
We say that an MDP $M$ is \emph{linear Bellman complete}. if,
for each $h \in [H]$, there exists a parameter $\theta'\in \RR^d$ such that
\[
\sup_{\theta \in \RR^d}
\;\sup_{(s,a) \in \mathcal{S} \times \mathcal{A}}
\; \Bigl | \Bigl\langle \phi_h(s,a), \,\theta' \Bigr\rangle
\;-\;
\mathbb{E}_{s' \sim \mathcal{P}_h(\cdot \mid s,a)}
\Bigl[
  r_h(s,a) 
  \;+\;
  \max_{a' \in \mathcal{A}} 
      \langle \phi_{h+1}(s', a'), \,\theta \rangle
\Bigr] \Bigr | \; = \; 0.
\]
\end{assumption}

\begin{figure}[h]
    \centering
    \includegraphics[width=0.5\linewidth]{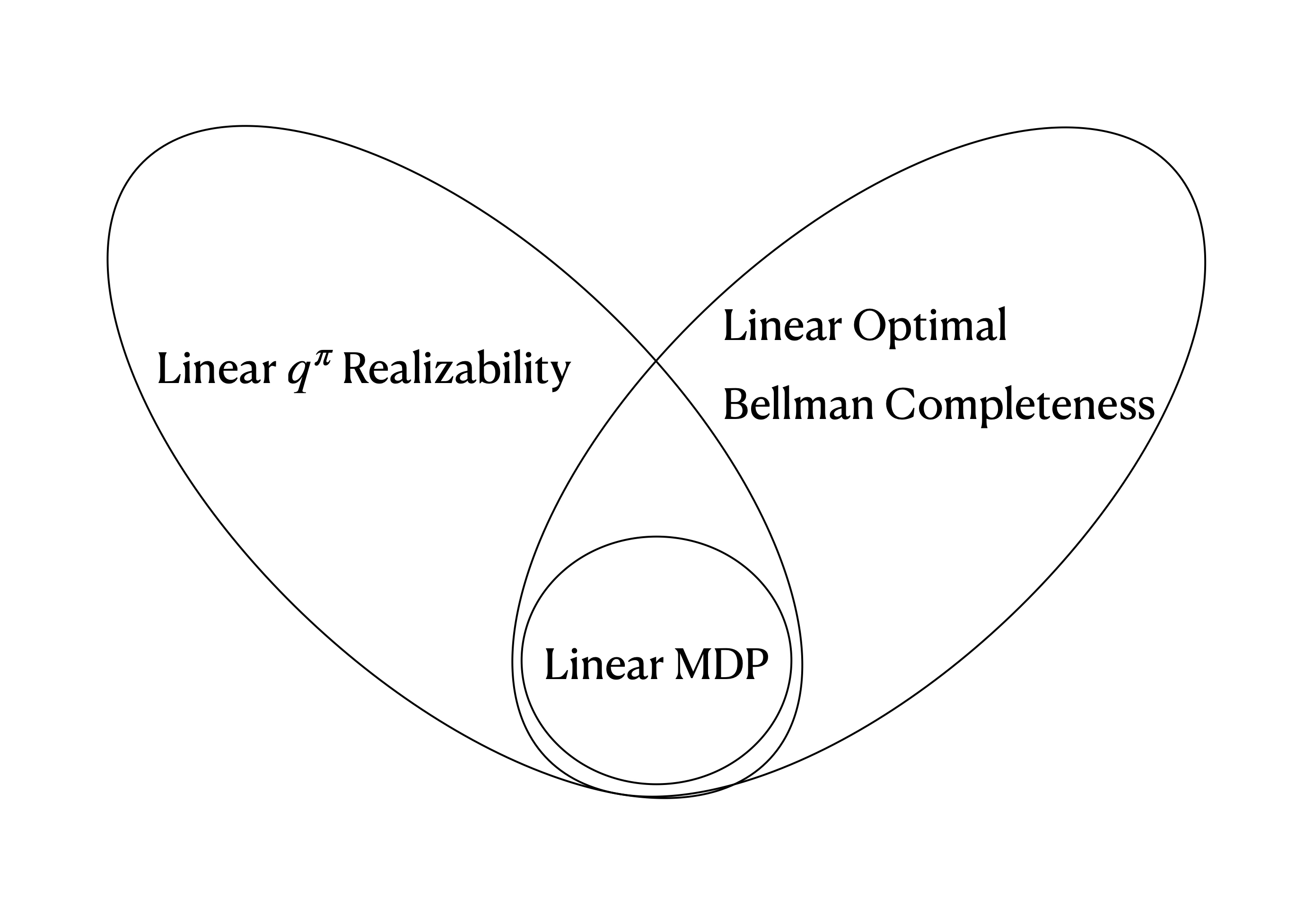}
    \caption{In the linear approximation setting, linear Bellman completeness and linear all-policy realizability neither imply nor dominate each other. Their intersection forms the strongest assumption among these, called the linear MDP assumption.}
    \label{fig:linear_MDP}
\end{figure}

Notice that adding more functions to the hypothesis class preserves realizability since value functions remain representable. However, it can break Bellman completeness by introducing functions whose Bellman updates no longer stay within the class. Consequently, requiring Bellman completeness can be less desirable, as it restricts how the function class can be expanded. In the linear approximation setting, linear Bellman completeness and linear all-policy realizability neither imply nor dominate each other shown in Figure \ref{fig:linear_MDP}. Their intersection forms the strongest assumption among these, providing a clean theoretical baseline for developing and analyzing algorithms. However, once these simpler settings are well understood, the field typically moves on to weaker assumptions that admit broader applicability.

\begin{assumption}[Linear MDP]
\label{assump:linear-mdp}
We say an MDP is a \emph{linear MDP} if there exist $d$ unknown (signed) measures $\mu_h = (\mu_h^{(1)}, \dots, \mu_h^{(d)})$ over $\mathcal{S}$ and an unknown vector $\theta_h \in \mathbb{R}^d$ such that
\[
    \mathcal{P}_h(s_{h+1}' \mid s_h, a_h) = \langle \phi(s_h,a_h), \mu_h(s'_{h+1}) \rangle,
    \quad
    \mathbb{E} \bigl[ r_h(s_h, a_h) \bigr] = \langle \phi(s_h,a_h), \theta_h \rangle
\]
for all $(s_h, a_h,s'_{h+1}) \in \mathcal{S}_h \times \mathcal{A}_h \times \mathcal{S}_{h+1}$ at each step $h \in [H]$. 

\noindent
Here, we assume 
\[
    \max \bigl\{ \|\mu_h(\mathcal{S}_h)\|, \|\theta_h\| \bigr\} \leq \sqrt{d}
\]
at each step $h \in [H]$, where, with an abuse of notation, we define
\[
    \|\mu_h(\mathcal{S}_h)\| = \int_{\mathcal{S}_h} \|\mu_h(s)\| \, ds.
\]
\end{assumption}

\begin{figure}[h]
    \vspace{-4mm}
    \centering
    \includegraphics[width=0.7\linewidth]{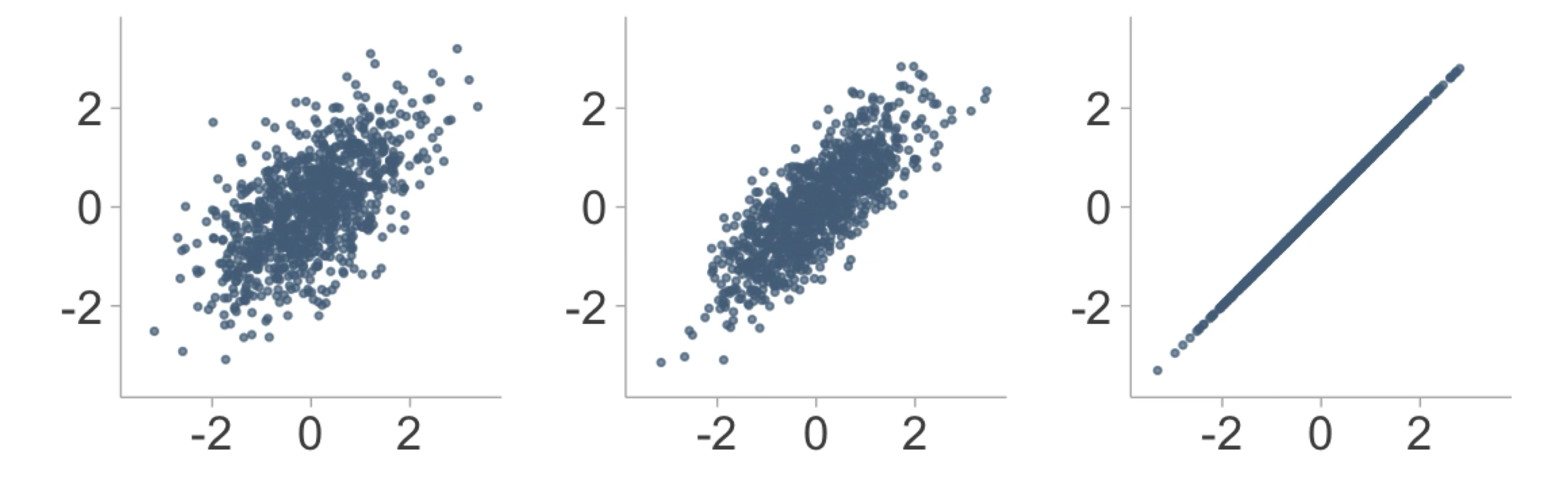}
    \vspace{-2mm}
    \caption{Each dot represents a two-dimensional feature vector. The data spreads and covers all dimensions in the left figure, while in the right figure, the data only gathers around a line without covering the perpendicular dimension. In the middle figure, the less covered dimension corresponds to a smaller eigenvalue.}
    \label{fig:feature_span}
    \vspace{-3mm}
\end{figure}

\subsection{Coverage}
Coverage assumptions constrain the dataset or the sampling distribution, requiring it to cover key state-action pairs with sufficiently high probability. A commonly used assumption is feature coverage, which states that the dataset should include state-action pairs whose features span the $d$-dimensional space as shown in Figure \ref{fig:feature_span}. This is equivalent to requiring the feature matrix, where each row represents the feature vector of a state, to be full rank.

In supervised learning, this assumption, combined with target realizability, is sufficient to guarantee learning with polynomially many samples.

\begin{assumption}[Feature Coverage]
\label{assump:feature-coverage}
Assume that for each \(h \in [H]\), the data distributions 
\(\mu_h\) satisfy the following minimum eigenvalue condition:
\[
    \sigma_{\min} \left( \mathbb{E}_{(s_h,a_h) \sim \mu_h} \big[ \phi(s_h,a_h) \phi(s_h,a_h)^\top \big] \right) \ge C_{\text{span}}.
\]
\end{assumption}

\begin{figure}[ht]
    \centering
    \includegraphics[width=0.3\linewidth]{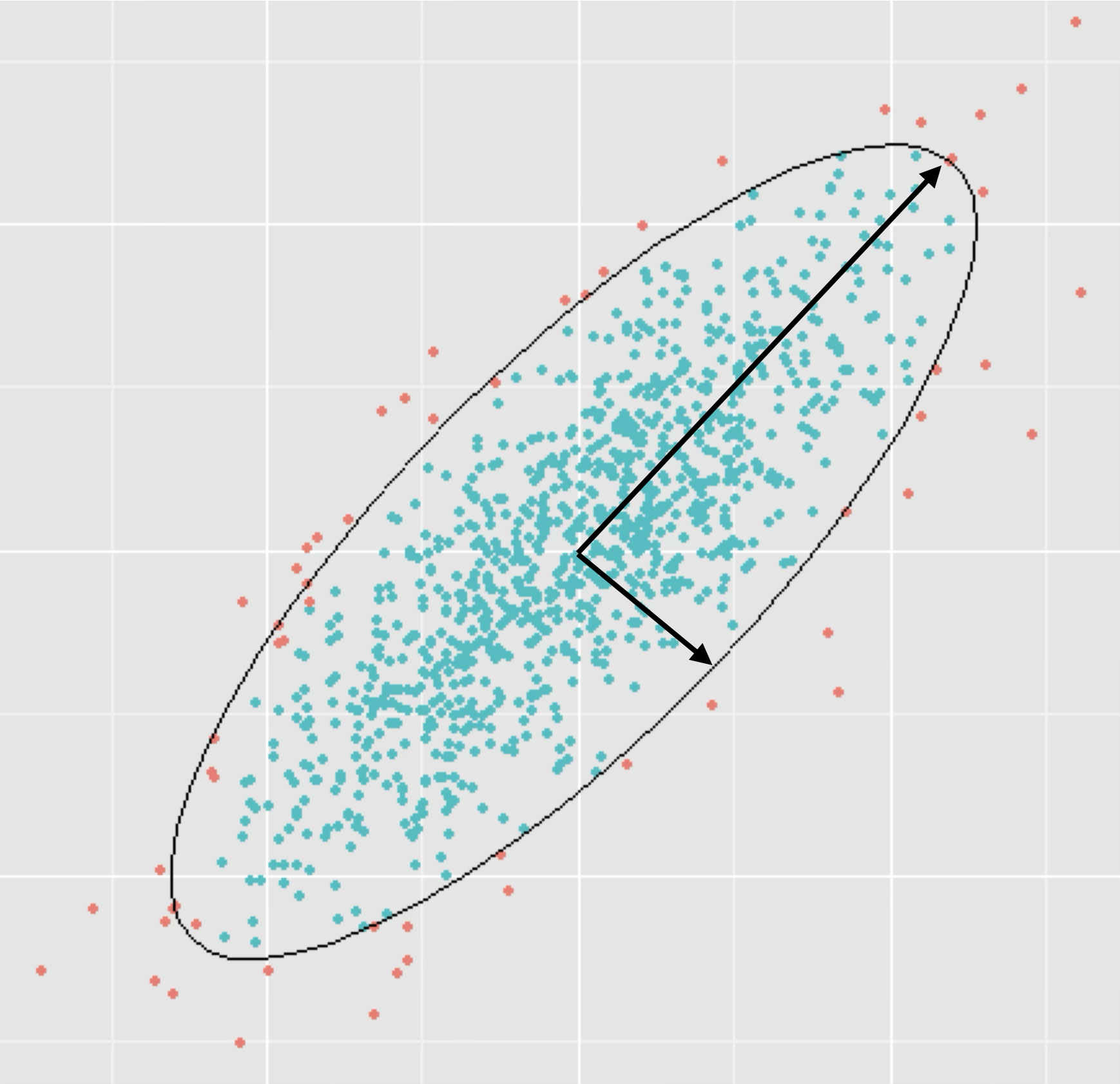}
    \caption{The eigenvectors of the covariance matrix are approximately illustrated in the figure. The eigenvector corresponding to the largest eigenvalue represents the direction of the greatest variance in the data. The second eigenvector, which is perpendicular to the first, indicates the direction of the second-largest variance.}
    \label{fig:min_eigenval}
\end{figure}

$\mathbb{E}_{(s_h,a_h) \sim \mu_h} \big[ \phi(s_h,a_h) \phi(s_h,a_h)^\top \big] \in \RR^{d\times d}$ represents a covariance matrix and $\sigma_{\min}$ denotes the smallest eigenvalue of a matrix. The eigenvector of the largest eigenvalue corresponds to the direction where data has the largest variance, as shown in Figure \ref{fig:min_eigenval}. The largest eigenvalue roughly measures the variance in this direction.
If the smallest eigenvalue is large, then the data covers all directions in feature space.

Reinforcement learning requires a different coverage condition from supervised learning. It is important to know the path leading toward the rewards and which action to take at each state, instead of only estimating value functions with small errors. So a more natural assumption is to ensure coverage of all state-action pairs that are reachable under all policies. This means that the dataset includes enough information to decide the order of actions and pick the optimal action.

\begin{assumption}[Concentrability - All-Policy Coverage]
\label{assump:all-coverage}
This condition asserts that there exists a constant $C_{\text{conc}} < \infty$ such that for all policy $\pi$ and step $h$,
\[
\sup_{\pi,h} \left \| \frac{P^\pi_h}{\mu} \right\|_{\infty} := \sup_{(s,a) \in \mathcal{S} \times \mathcal{A}} \left\{ \frac{\PP^\pi_h(S_h=s,A_h=a)}{\mu(s,a)} \right\} \leq C_{\text{conc}},
\]
where $P^\pi_h$ represents the state-action visitation distribution $\PP^\pi_h(S_h=s,A_h=a)$.
\end{assumption}

This concentrability coefficient $C_{\text{conc}}$ measures the maximum discrepancy between the data sampling distribution $\mu$ and the state-action visitation distributions induced by any policy $\pi$. A small coefficient ensures high probabilities to cover trajectories sampled under any policy $\pi$.

All-policy coverage is fairly strong and is difficult to satisfy. In tasks such as Go, almost all state-action pairs are accessible under some policy. Setting the data collection policy to be uniform over all state-action pairs can satisfy the assumption. However, this causes the concentrability coefficient to depend on the size of the state space, resulting in an enormous data requirement. Therefore, we adopt a single-policy coverage assumption instead.

\begin{assumption}[Single-Policy Coverage]
\label{assump:single-coverage}
This condition asserts that there exists a constant $C_{\text{conc},\,\pi} < \infty$ such that for some policy $\pi$,
\[
 \sup_{h \in [H]}\sup_{(s,a)\in \mathcal{S} \times \mathcal{A}} \left\{\frac{\mathbb{P}^{\pi}_h(s,a)}{\mu_h(s,a)} \right\} \leq C_{\text{conc},\,\pi} \,. 
\]
\end{assumption}

\begin{table}[t]
    \hspace{-10mm}
    \begin{tabular}{|c|c|c|c|}
        \toprule
         \multicolumn{3}{|c|}{Conditions} & Tractability\\
         \midrule
         Representation & Data Coverage & Sampling  & \makecell{Tractability with \\ $Poly(d,H,|\cA|,\frac{1}{\epsilon})$ Data} \\
         \midrule
         \makecell{Bellman Completeness + \\$q^*$-Realizability} & NaN & NaN & No \citep{chen2019information}\\
         All-Policy Realizability & NaN & Policy-Induced & No \citep{xiao2022curse}\\
         \midrule
         All-Policy Realizability & Feature Coverage & NaN & No \citep{wang2021statistical}\\
         All-Policy Realizability & All-Policy Coverage & NaN & No \citep{foster2022offline}\\
         All-Policy Realizability & All-Policy Coverage & Policy-Induced & Yes 
         \citep{tkachuk2024trajectory}\\
         Bellman Completeness & All-Policy Coverage & NaN & Yes \citep{antos2008learning}\\
         \midrule
         Linear MDP & Single-Policy Coverage & Policy-Induced & Yes \citep{jin2021pessimism}
         \\
         \makecell{Bellman Completeness $+$ \\ Linear Deterministic Rewards} & Single-Policy Coverage & NaN & Yes \citep{golowich2024role} \\
         Single-Policy Realizability & Single-Policy Coverage & Policy-Induced & No \citep{jia2024offline} \\
         \bottomrule
    \end{tabular}
    \caption{This table lists different choices of conditions and whether an algorithm exists to solve all tasks satisfying given conditions in a polynomial amount of data.}
    \label{tab:condition}
\end{table}

\section{Hardness of Offline RL}
\label{sec:hardness}
The goal is to learn a near-optimal policy such that $\mathrm{SubOpt}\bigl(\hat{\pi}; x\bigr) \le \epsilon$ with high probability. However, prior results \citep{chen2019information, xiao2022curse} demonstrate that, regardless of whether we assume all-policy realizability or Bellman completeness, the task is not solvable within a polynomial number of samples $O(d,H,|\cA|,\frac{1}{\epsilon})$ without dependence on the state space size $|\cS|$, if no coverage assumption is made. This highlights the fundamental challenge of offline RL—without sufficient data coverage, even strong function class assumptions cannot guarantee efficient learning.

We first adopt the all-policy realizability assumption, ensuring that all policies can be accurately evaluated by some function within the function class such that an algorithm can update and evaluate policies iteratively. A natural next step is to consider feature coverage, an assumption that is typically sufficient in supervised learning settings. However, as shown in Theorem \ref{thm:feature+all-realizability}, this assumption alone is insufficient for offline RL \citep{wang2021statistical}. A proof for the infinite horizon setting is also developed \citep{amortila2020variant}.

\begin{theorem} [\citet{wang2021statistical}]
\label{thm:feature+all-realizability}
Suppose a dataset satisfying Assumption \ref{assump:feature-coverage} is given. There exists an MDP and feature representation satisfying Assumption \ref{assump:all-realizability}, such that any algorithm requires $\Omega((d/2)^H)$ samples to output a policy $\hat{\pi}$, giving $\mathrm{SubOpt}\bigl(\hat{\pi}; x\bigr) \le 0.5$ with probability at least $0.9$.
\end{theorem}

The theorem introduces a lower bound of $\Omega((d/2)^H)$ samples, indicating that the sample complexity grows exponentially with the horizon $H$. This result underscores the need for stronger coverage or representation conditions to achieve efficient offline RL.

Such a bound highlights the inherent hardness of offline RL. First, current value estimation uses bootstrapping \citep{sutton1998introduction}, which relies on estimated values for the state-action pairs at later steps. Instead of observing the true value functions, learning relies on bootstrapping targets, which introduce two sources of estimation bias. First, noise from the stochasticity of rewards and transitions leads to statistical errors. Second, the bootstrapping target may not be realizable \citep{weisz2023online}, resulting in approximation errors. Specifically, the bootstrapping target is given by
\[
    r(s, a) + \max_{a'} \langle \phi(s', a'),\theta \rangle,
\]
which may not be representable for all state-action pairs and all parameters $\theta\in\RR^d$. Note that, unlike stochastic noise, the bias caused by unrealizability cannot be reduced by increasing the amount of data. \looseness=-1

\begin{figure}[h]
    \centering
    \includegraphics[width=0.5\linewidth]{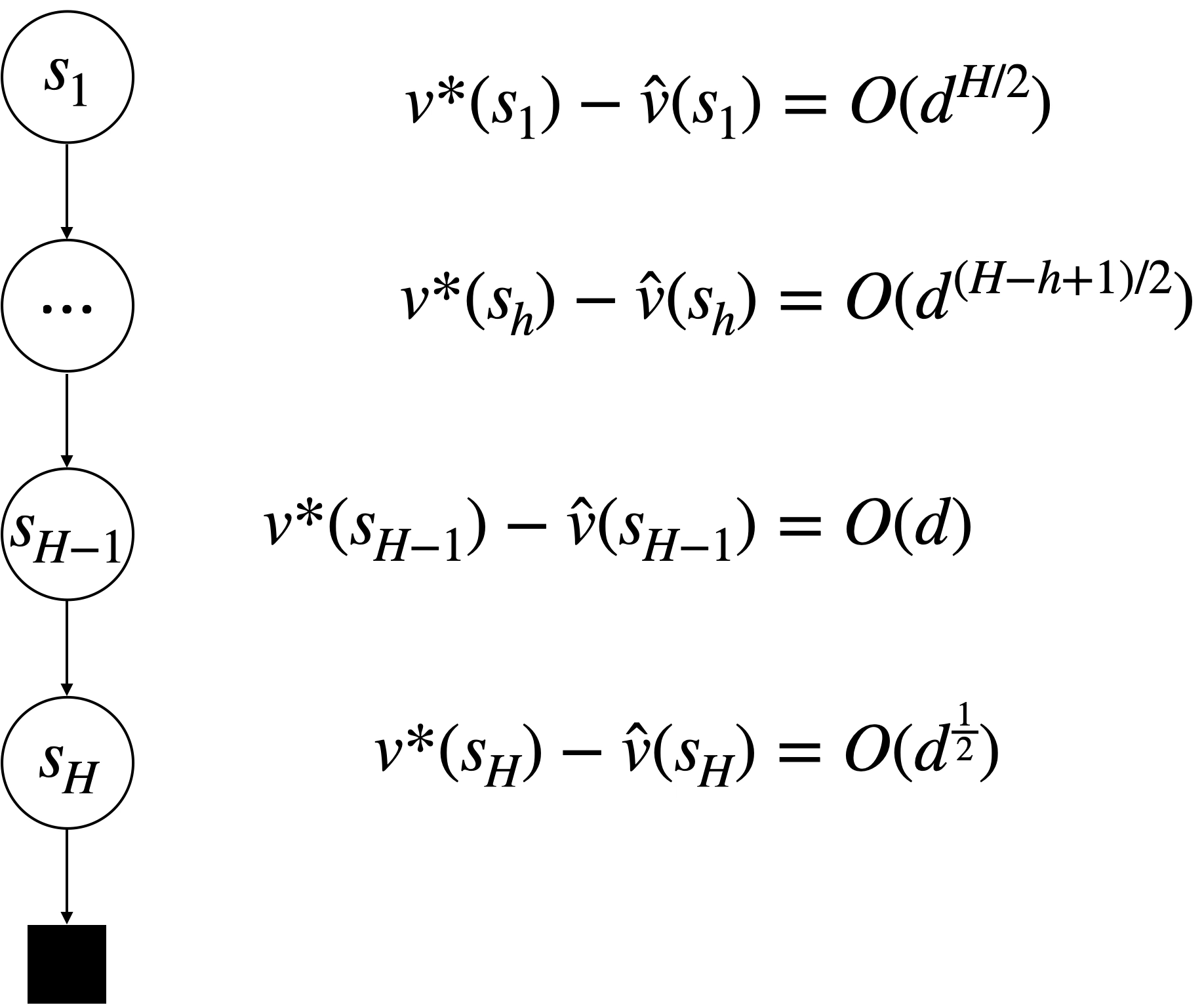}
    \caption{Errors in the estimation of bootstrapping targets can accumulate and amplify exponentially over the horizon, from $O(\sqrt{d}$ at step $H$ to $O(d^{\frac{H}{2}})$ at the initial state.}
    \label{fig:bias}
\end{figure}

These errors in the estimation of bootstrapping targets can accumulate and
amplify exponentially over the horizon, from $O(\sqrt{d})$ at step $H$ to $O(d^{\frac{H}{2}})$ at the initial state, as shown in Figure \ref{fig:bias}. 

Another major challenge is that the dataset must cover the path leading to rewards by following the optimal policy. The difficulty in reinforcement learning is not just about estimating value functions with small errors. The goal is to identify which actions lead to higher returns. With only all-policy realizability, there is no generalization across policies: knowing the return of executing policy $\pi_A$ does not provide any information about the return of another policy $\pi_B$. Without data that captures the trajectory leading to high rewards, algorithms cannot infer the optimal policy. As a result, no algorithm can learn a policy that outperforms the behaviours present in the dataset without further interactions.

Due to this issue, researchers focus on tasks which satisfy an alternative coverage assumption known as concentrability, which directly assumes sufficient data coverage over trajectories generated by policies. This assumption ensures coverage of all state-action pairs that are reachable under some policies. However, recent results \citep{foster2022offline} demonstrate that even under this assumption, the task remains unsolvable within a polynomial amount of data, highlighting the need for even stronger assumptions or novel algorithmic techniques to make offline RL feasible.

\begin{theorem} [\citet{foster2022offline} ]
\label{thm:all-coverage+all-realizability}
Suppose a data distribution $\mu =(\mu_h)_{h=1}^H$ satisfying Assumption \ref{assump:all-coverage} is given with concentrability coefficient $C_{\text{conc}} \leq 16$. For any dataset collected under this distribution, there exists an MDP and a general function class $\mathcal{F}$ satisfying Assumption \ref{assump:all-realizability} such that any algorithm requires at least $c \cdot S^{1/3}$ samples to output a policy $\hat{\pi}$ with
\[
v^{\star}_1(x) - \mathbb{E}_{\cD}[v^{\hat{\pi}}_1(x)] \le \frac{c'}{(1 - \gamma)},
\]
where $c$ and $c'$ are absolute numerical constants.
\end{theorem}

The challenge arises because the dataset may contain spurious data that degrades performance. Some state-action pairs included in the dataset may not be accessible in the environment, yet they are still present in the offline data. For example, consider a cleaning robot: if the dataset contains instances of the robot being placed on a shelf — an infeasible scenario during real execution — the learning algorithm has no way of identifying these as inaccessible. As a result, these spurious data disrupt value estimation, leading to inaccurate policy learning.

To address this issue, researchers focus on policy-induced data under Assumptions \ref{assump:all-realizability} and \ref{assump:all-coverage}, ensuring both all-policy realizability and sufficient coverage. The implications of this approach are discussed in the next section.

\paragraph{Key Takeaways.}
The first challenge of offline reinforcement learning (RL) is that the true value function is not directly observed, and learning typically relies on bootstrapping. However, \textbf{bias in estimating the bootstrapping target can accumulate and amplify exponentially along the horizon}. 

One source of bias arises when the bootstrapping target is not representable, leading to approximation errors. Unlike stochastic noise from rewards and transitions, this type of bias cannot be reduced by collecting more data and needs extra techniques to control it.
A proposed approach is to select a representable $n$-step return as the bootstrapping target \citep{tkachuk2024trajectory}. A more practical alternative is to use the $\lambda$-return \citep{sutton1998introduction}, which reduces dependence on a single step and helps mitigate bias from unrealizability.

The second challenge is that \textbf{there is no value function generalization between policies}.
The realizability condition of value functions alone does not allow generalization across policies: knowing the return of executing policy $\pi_A$ does not provide any information about the return of another policy $\pi_B$. Without data that captures the trajectory leading to high rewards, algorithms cannot infer the optimal policy. As a result, \textbf{no algorithm can learn a policy that outperforms the behaviours presented in the dataset without further interactions with the environment}.

The third challenge comes from \textbf{spurious data}, which includes transitions messing up the value estimation but cannot be distinguished by the agent. A systematic solution is to use policy-induced data to avoid including any inaccessible but misleading data.

\begin{table}[t]
    \hspace{-12mm}
    \begin{tabular}{|c|c|c|c|}
        \toprule
         \multicolumn{3}{|c|}{Conditions} & Upper Bound\\
         \midrule
         Representation & Data Coverage & Sampling  & \makecell{Corresponding SubOpt -- \\Initial State Value Difference Bound} \\
         \midrule
         All-Policy Realizability & All-Policy & Policy-Induced &   $\tilde{\Theta}(C^2 d^2H^\frac{7}{2} n^{-\frac{1}{2}})$   \citep{tkachuk2024trajectory}\\
         Bellman Completeness & All-Policy & NaN & \makecell{$\tilde{O}(C n^{-\frac{1}{2}}(1-\gamma)^{-4})$ \\ \citep{chen2019information}}\\
         \midrule
         Linear MDP & Single-Policy & Policy-Induced & $\tilde{O}(C d^2H^2 K^{-\frac{1}{2}})$ \citep{jin2021pessimism}
         \\
         \makecell{Bellman Completeness $+$ \\ Linear Deterministic Rewards} & Single-Policy & NaN & \makecell{$\tilde{O}((C+H) d H n^{-\frac{1}{2}} + \epsilon_{\textrm{actor}})$ \\ \citep{golowich2024role}} \\
         \bottomrule
    \end{tabular}
    \caption{This table lists the sample complexity under different sufficient conditions. $C$ is the corresponding concentrability coefficient, $d$ is the linear function approximation feature dimension, $H$ is the horizon, $n$ is the number of samples, $K$ is the number of trajectories and $\epsilon_{\textrm{actor}}$ denotes the optimization error in actor updates. Also, $\gamma$ refers to the discount factor and implies the result for the infinite horizon setting.}
    \label{tab:condition}
\end{table}

\section{Sufficient Conditions}
\label{sec:sufficient}
This section first presents the sufficient condition under the all-policy coverage assumption and introduces the techniques used to establish polynomial sample efficiency under these conditions. Next, we relax this impractical coverage assumption by considering a setting where only state-action pairs accessible by a single policy need to be covered. This condition is formally stated in Assumption~\ref{assump:single-coverage} and is referred to as the single-policy coverage assumption.

\subsection*{All-Policy Coverage}

As discussed in the last section, even under strong representation and coverage assumptions, such as all-policy realizability and coverage, not all tasks are solvable with a polynomial number of samples. The fundamental issue is that the dataset may contain spurious data, meaning state-action pairs that are not accessible by any policy in the environment. These extraneous data points can mislead value estimation and degrade policy performance. To mitigate this problem, we introduce our first sufficient conditions, which imposes an additional policy-induced assumption: the dataset must contain complete trajectories generated by a valid policy in the environment. By ensuring that all transitions in the dataset are achievable under some policy, we eliminate inaccessible state-action pairs and prevent them from corrupting value estimation. This assumption strengthens the generalization capability of offline RL, as learning is now constrained to relevant transitions.

\begin{theorem}[\citep{tkachuk2024trajectory}]
\label{thm:vlad}
Let $\epsilon>0$. Consider a dataset of complete trajectories collected under a data distribution satisfying \ref{assump:all-coverage}, and an MDP and feature representation satisfying Assumption \ref{assump:all-realizability}.

There exists an algorithm which requires $n = \tilde{\Theta} \left( C_{\text{conc}}^4 H^7 d^4 / \epsilon^2 \right)$ number of trajectories such that with probability at least $1 - \delta$, the output policy $\hat{\pi}$
\[
\mathrm{SubOpt}\bigl(\hat{\pi}; x\bigr)\leq \epsilon,
\]
where $\tilde{\mathcal{O}}$ is equivalent to $\mathcal{O}$ in big-Oh notation without polylogarithmic factors. 
\end{theorem}

The use of complete trajectories helps eliminate spurious data while simultaneously addressing the unrealizable one-step bootstrapping target issue by leveraging $n$-step returns. Theorem \ref{thm:vlad} uses a computationally inefficient algorithm, which automatically determines the appropriate step number for return estimation, ensuring that only realizable bootstrapping targets using multi-step returns are used for each state. A similar approach has been developed and adopted by some algorithms, known as $\lambda$-return. A $n$-step return is defined as
\begin{equation}
    G_{t:t+n}(s) = \sum_{h=t}^{t+n-1} R_h(s_h,a_h) + v^{\pi}_{t+n}(s_{t+n}).
\end{equation}

$\lambda$-return computes the bootstrapping target as a weighted sum of all-step returns, given by
\begin{equation}
    G_t^{\lambda}(s) = \frac{1-\lambda}{1-\lambda^{H-t}}\sum_{n=1}^{H-t} \lambda^{n-1} G_{t:t+n}(s),
\end{equation}
where setting $\lambda=0$ results in a one-step return and setting $\lambda=1$ gives a Monte-Carlo return estimated by the sum of rewards until termination.

An important open question is how to design a state-dependent $\lambda$ that dynamically weights realizable steps more heavily, reducing the bias of bootstrapping. This remains an area for future exploration.

The all-policy coverage assumption is difficult to satisfy in practice. In certain tasks, such as Go, almost all state-action pairs are accessible under some policy. However, requiring uniform coverage over all state-action pairs imposes an excessively strong assumption. A natural way to satisfy this assumption is to use a uniform data collection policy over all state-action pairs. Unfortunately, this approach results in a concentrability coefficient that scales with the size of the state space, leading to an impractically large data requirement.

\subsection*{Single-Policy Coverage}

Let's start weakening the all-policy coverage assumption. As a starting point, we adopt the strongest representation assumption, that is, linear MDPs, defined in Assumption~\ref{assump:linear-mdp}. The single-policy coverage assumption assumes the sampling distribution to cover state-action pairs accessible under a single policy $\pi$. This assumption describes the practical situation. For instance, in autonomous driving, the dataset is typically collected from human drivers, limiting coverage to the behaviours demonstrated by humans \citep{hester2018deep}.

As discussed in the previous section, the learned policy cannot be guaranteed to outperform the best policy contained in the dataset. With only all-policy realizability, there is no generalization across policies, meaning the algorithm can only perform comparably to the demonstrated behaviours in the dataset. Consequently, the learning objective must change: instead of aiming to approximate the optimal policy, the goal is now to learn a policy that is comparable to the behaviour policy $\pi$.

\begin{remark}
    Under the single-policy assumption, where a behaviour policy $\pi$ is used to collect data (as stated in Assumption~\ref{assump:single-coverage}), the goal of offline reinforcement learning is to learn a policy $\hat{\pi}$ such that, with high probability,
    \[
    \left| v_1^\pi(x) - v_1^{\hat{\pi}}(x) \right| \leq \epsilon,
    \]
    where the initial state is \( s_1 = x \).
\end{remark}

However, partial coverage of the state-action space can lead to overestimated extrapolation, where the algorithm assigns unrealistically high-value estimates to poorly represented or unseen state-action pairs. Since these data are not covered enough, the estimation bias is higher due to the variance. Then, the agent may overcommit to exploratory actions with insufficient support in the dataset. Even if the optimal policy is covered by the dataset, the agent can deviate from it and degrade overall performance.

To address this issue, pessimism has been developed to prevent the agent from choosing less-visited state-action pairs.
Pessimism penalizes the value functions of state-action pairs that are not covered well by the dataset. An uncertainty term $U(s,a)$ is reduced from the value estimation
\begin{equation}
    q_{LCB}(s,a) = \hat{q}(s,a) - \lambda U(s,a),
\end{equation}
where $\lambda$ is a weighting coefficient of the uncertainty and can be considered as one.
This uncertainty term is roughly proportional to $\sqrt{\frac{1}{\# (s,a) \text{ in the dataset}}}$ in the tabular setting. Thus, the pessimism reduces value estimations of less-visited state-action pairs and constrains the agent to state-actions well-covered by the dataset.

When using a linear MDP, the bootstrapping target, after subtracting this uncertainty term, remains linearly realizable. As a result, we gain the following theorem.

\begin{theorem}[\citep{jin2021pessimism}]
Consider a linear MDP (Assumption \ref{assump:linear-mdp} is satisfied) and a dataset of $K$ trajectories covering a policy $\pi$ with a concentrability coefficient $C$ (Assumption \ref{assump:single-coverage} is satisfied).
Let $\lambda = 1$ and $\beta = c \cdot dH \sqrt{\zeta}$, where $\zeta = \log(2dHK / \delta)$.  
There exists an algorithm, with probability exceeding $1 - \delta$, we have:
\begin{align}
    v^\pi(x) - v^{\hat{\pi}}(x) &\leq 2\beta \sum_{h=1}^{H} \mathbb{E}_{\pi} \left[ \left( \phi(s_h, a_h)^\top \Lambda_h^{-1} \phi(s_h, a_h) \right)^{1/2} \Big| s_1 = x \right] \\
    & \le 2 \beta C_{\text{conc},\,\pi}  O(\frac{d}{\sqrt{n}}).
\end{align}
\end{theorem}

Next, we remove the linear MDP assumption and adopt a weaker yet still convenient assumption—Bellman completeness. Several works attempt to simplify the condition following the actor-critic framework and leverage variants of the Bellman completeness condition \citep{xie2021bellman,zanette2021provable}. Finally, \cite{golowich2024role} establish a proof under the Bellman completeness condition by focusing on a special class of policies, called linear perturbed policies. The Bellman completeness assumption ensures that the original bootstrapping target, $ r(s,a) + \max_{a'} \hat{q}(s',a')$, is representable within the function class. However, when incorporating an uncertainty term reduction, the resulting target may no longer be linearly representable. This necessitates finding an alternative approach to effectively embed pessimism, introduced by \cite{zanette2020learning}.

To incorporate pessimism, researchers propose using a lower bound of the estimated Q-values, $q^\pi$. 
Specifically, they consider a small ball centred around the original estimate, carefully controlling its size to ensure that, with high probability, the true value lies within the ball. 
Any estimate within this region represents a Q-value under an MDP that is close to the true one.
Pessimism, in this context, requires designing an algorithm that performs well under the worst-case estimate within this uncertainty set. 
That is, for any estimation within this neighbourhood, the performance difference between the corresponding policy and the target policy remains small.
As a result, the adopted approach selects the lowest possible Q-value estimate within the ball, ensuring a conservative policy.

The detailed algorithm is presented below. $\hat{\theta}$ is the original parameter learnt by least square TD to evaluate Q-values of policy $\pi$. But pessimism chooses another parameter $\theta = \xi + \hat{\theta}$, which gives the lowest Q-value estimation in a certain region around $\hat{\theta}$. The parameter is updated as
\begin{align}
    (\xi^\pi, \theta^\pi) & =:\arg\min_{\xi \in (\mathbb{R}^d)^H, \, \theta \in (\mathbb{R}^d)^H} \sum_{a \in \mathcal{A}} \pi_1(a \mid s_1) \langle \phi_1(s_1, a), \theta_1 \rangle
\end{align}
with the terminal condition \( \theta_{H+1} = 0 \), and subject to the constraints for $h \in [H]$,
\begin{align}
    \theta_h &= \xi_h + \hat{\theta}_h, \\
    \|\xi_h\|^2 &\leq \alpha_h^2, \quad \|\theta_h\|^2_2 \leq \beta_h^2. 
\end{align}
Here, $\langle \phi_1(s_1, a), \theta_1 \rangle$ is the estimated Q-value and expectation over all actions gives the state value for the initial state $\hat{v}_1(s_1)$. Parameters which give the lowest initial state value are chosen to include pessimism.

Next, we present the upper bound of sample complexity under the Bellman completeness and single-policy coverage, when injecting pessimism through the lower bound of value estimation. This result requires an extra assumption on rewards, shown below.

\begin{assumption}
For any step $h$, the reward function is deterministic and satisfies
\[
    r_h(s,a) = \langle \phi_{h}(s,a), \theta_{h}^r \rangle.
\]
\label{assump:linear_reward}
\end{assumption}

\begin{theorem}[Informal \citep{golowich2024role}]
Consider an MDP and feature representing which satisfied Assumption \ref{assump:low-inherent-bellman-error} and a dataset covering a policy $\pi$ with a concentrability coefficient $C_{\textrm{conc}\, , \pi}$ (Assumption \ref{assump:single-coverage} is satisfied).
There exists an algorithm running in polynomial time, which outputs a policy $\hat{\pi}$ at random such that for any policy $\pi$,
with probability exceeding $1 - \delta$, we have:
\begin{align}
    v^\pi(x) - v^{\hat{\pi}}(x) &\leq \tilde{O} \left( 
\frac{ B H d }
{\sqrt{n}}
\right) \cdot (H + C_{\mathcal{D}, \pi}) + \epsilon_{\text{final}}.
\end{align}
\end{theorem}

There is no existing theoretical work on the solvability of offline RL under the combination of all-policy realizability, single-policy coverage, and policy-induced data, which the author suspects to be sufficient using the same $n$-step return as our first sufficient result from Tkachuk et.\ al (2024). However, it has been proven that switching to single-policy realizability plus others is insufficient \citep{jia2024offline}.
Intuitively, algorithms such as policy iteration or actor-critic require the ability to evaluate any policy encountered during the learning process. Without the capacity to accurately represent the value function of each policy, bias can accumulate, ultimately leading to suboptimal learning outcomes.

\subsection{Practical Implementation of Pessimism}
We have introduced two ways of embedding pessimism: reduction of an uncertainty term from the value estimation or the usage of a lower bound of the value estimation. These two ways are also adopted in practice. 

Clipped double Q-learning \citep{fujimoto2018addressing} is widely used in actor-critic and policy iteration algorithms. This approach employs two networks $\hat{q}_{\theta_{1}}$ and $\hat{q}_{\theta_{2}}$ that share the same target, relying on the minimum value estimation between them for bootstrapping. The policy is evaluated on the first network as follows: $\pi(s') = \arg\max \hat{q}_{\theta_1}(s',a')$, with the target defined as
\begin{equation}
    r(s,a) + \gamma \min_{i \in \{1,2\}} \hat{q}_{\theta_{i}}(s',\pi(s')).
\end{equation}
This method aims to alleviate the bias of taking maximum over random variables instead of their means and thus reduce the overestimation of extrapolation. Building on this idea, several algorithms also take the minimum across multiple Q-networks \citep{van2016deep, lan2020maxmin}.

It turns out that 
\begin{equation}
    \min_{i \in \{1,2\}} \hat{q}_{\theta_i}(s,a) = \frac{\hat{q}_{\theta_1}(s,a)+\hat{q}_{\theta_2}(s,a)}{2} -\frac{|\hat{q}_{\theta_1}(s,a)-\hat{q}_{\theta_2}(s,a)|}{2},
\end{equation}
where $\frac{|\hat{q}_{\theta_1}(s,a)-\hat{q}_{\theta_2}(s,a)|}{2}$ roughly measures the variation of Q-value estimations and serves as the uncertainty reduction.

Conservative Q-learning \citep{kumar2020conservative} is one of the best-performing offline RL algorithms and follows the second way of pessimism: it directly regularizes Q-value estimations to keep the value low. Its algorithm follows value iteration, and the update rule at iteration $k+1$ is
\begin{equation}
\hat{q}^{k+1} \leftarrow \arg\min_{Q} \alpha \mathbb{E}_{s \sim \mathcal{D}, a \sim \mu(a|s)} [Q(s,a)] + \frac{1}{2} \mathbb{E}_{s,a \sim \mathcal{D}} \left[ \left( Q(s,a) - \mathcal{B}^{\pi} \hat{q}^{k}(s,a) \right)^2 \right].
\end{equation}
The update rule does not consider a high-probability small error region around the least-square temporal difference (LSTD) learning fixed point \citep{bradtke1996linear}, as the algorithm with theoretical analysis requires. That means the regularization may cause a large deviation during value estimation and gives a policy with a performance guarantee. 

\paragraph{Key Takeaways.}
Current sufficient conditions are the combination of Bellman completeness and single policy coverage. These are not just assumptions used for theoretical convenience; they also represent the essential conditions for our algorithm to function effectively. To extend our approach to a broader range of tasks under weaker conditions, novel techniques are required—rather than endlessly tuning the model. Moreover, current results suggest that we need to control the estimation error of bootstrapping targets instead of value functions, which should be examined to ensure a strong enough representation ability of the used networks.

\cite{dong2020expressivity} have constructed MDPs with piecewise linear transition dynamics where the optimal $q^*$ function has a high-degree complexity, making precise learning infeasible. Since our survey does not cover model-based algorithms, we note that in such cases, where data is limited, approximating the model itself may be a more effective strategy.

When the dataset provides only partial coverage of state-action pairs, pessimism is often required to constrain the agent within the covered distribution. On the contrary, optimism promotes exploration in less-visited areas. Interestingly, clipped double-Q \citep{fujimoto2018addressing, chen2021randomized,cetin2023learning}, a pessimistic approach, is widely used in online RL. This raises the question: what prevents an online algorithm from succeeding without pessimism? It turns out that online algorithms are also suffering from the over-estimated extrapolation as well, which is benign for exploration, but causes the parameters of value networks to explode and the network to lose learning plasticity \citep{lyle2024disentangling}.
Other than pessimism, feature normalization and resetting also demonstrate the potential to replace traditional critic regularization techniques \citep{nauman2024overestimation, gallici2024simplifying, bhatt2024crossq}. 

\section{Low Adaptive Learning}
\label{sec:low-adaptivity}
Online learning collects novel transitions after each policy update and uses optimism to encourage exploration.
Offline learning, on the other hand, is given a fixed dataset and relies on pessimism to remain within the covered state-action space.
Between these two approaches is low-adaptive learning, where there are limited switches to collect data and the policy is updated multiple times using the collected dataset. 
Adaptivity refers to the frequency of policy changes.
This setting aligns better with the practical applications, where policy evaluation is required before execution for a safety check and policies continue to be updated to approach the optimal one with a novel batch of data collected. 

In this setting, two questions arise naturally: (1) Can we have guaranteed improvement at each iteration? (2) How many switches are required to learn a near-optimal policy? \cite{huang2024adaptive} analyze how to collect data for the next step to ensure guaranteed improvement of the learnt policy under pessimism. However, the paper shows a negative result, implying that either pessimism should not be adopted or per-switch improvement is not achievable.

\begin{table}[ht]
    \centering
    \begin{tabular}{c|c}
    \toprule
    Representation Condition & Lower and Upper Bounds of Switch Costs \\
    \midrule
    Tabular & $\Theta(HSA \log\log T)$ \citep{qiao2022sample} \\
    Linear MDP & $[\Omega(dH \log\log T), O(dH \log K)]$ \citep{qiao2022near,gao2021provably}\\
    \bottomrule
    \end{tabular}
    \caption{This table lists the required number of policy changes to achieve $\Tilde{O}(\sqrt{T \mathrm{poly}(d,H)})$ regret, of the same order as online learning. Here, $d$ is the linear function approximation feature dimension, $H$ is the horizon, $T$ is the number of samples, and $K$ is the number of trajectories.}
    \label{tab:placeholder}
\end{table}

Theoretical work suggests that, under certain conditions, achieving performance comparable to current online algorithms requires only $\log (T)$ policy switches (but more gradient updates) when collecting $T$ samples \citep{wang2021provably,gao2021provably}. The proof focuses on eliminating unqualified policies and deciding when to conduct policy change \citep{abbasi2011improved}. There has not been any algorithm that achieves $\Tilde{O}(\sqrt{T})$ regret under $O(\log\log(T))$ switching budgets and remains an open problem. Also, several works have started to explore beyond linear MDPs \citep{qiao2023logarithmic,zhao2024nearly}. Empirical researchers are also working on high replay ratio or high update-to-data ratio learning, with fewer policy changes and more reuse of data \citep{chen2021randomized,hiraoka2022dropout,hussing2024dissecting}. However, current algorithms follow the online RL setting and still require almost $O(n)$ switches. Theoretical results support us in exploring novel algorithms to more carefully choose whether to switch data collection policy.

\section{Related Works}
The tutorial does not cover all research directions in offline RL, but gives an introduction to theoretical results. Please refer to a survey on offline and low-adaptive RL for more details and settings beyond linear function approximation \citep{yin2025statistical}.

There is also a trend of works adding density ratio correction to directly solve the challenge of distribution shifts, but not covered in the tutorial \citep{zhan2022offline,chen2022offline,amortila2024harnessing,che2025avg}.
Meanwhile, the tutorial does not cover any instance-dependent result \citep{yin2021towards, chen2022offline,nguyen2023instance}, which leverages task-related parameters and usually gives a sharper bound than mini-max sample complexity.

This tutorial focuses on off-policy optimization; meanwhile, off-policy evaluation (OPE) is a starting point for offline RL, and several hardness results hold for OPE. The regression estimator has been shown to be mini-max optimal for tabular bandits \citep{li2015toward}. Fitted Q-evaluation (FQE) gives a nearly mini-max optimal finite-sample bound with high probability \citep{duan2020opeconvergence} under certain function class assumptions. The asymptotic distribution of the FQE value estimation error using linear function approximation is shown to follow a normal distribution, which can be used for confidence interval computation \citep{hao2021bootstrapping}. Furthermore, in infinite-horizon settings, variants of temporal difference algorithms \citep{sutton2009fast,sutton2016emphatic,che2024target} are developed. 

\section{Acknowledgment}
We sincerely appreciate the efforts of Ming Yin and Roshan Shariff in preparing this tutorial and the nice discussions with Csaba Szepesvari and Chenjun Xiao.

\clearpage
\bibliography{iclr2023_conference}
\bibliographystyle{iclr2023_conference}

\end{document}